\documentclass[conference]{IEEEtran}
\usepackage{cite}
\usepackage{amsmath,amssymb,amsfonts}
\usepackage{graphicx}
\usepackage{textcomp}
\usepackage{nicefrac}
\usepackage{pifont}
\usepackage{soul}
\usepackage[font=footnotesize,labelfont=bf]{caption}
\usepackage{enumitem}
\usepackage[caption=false, font=footnotesize]{subfig}
\usepackage[table,xcdraw]{xcolor}
\usepackage{titlesec}
\usepackage{anyfontsize}
\usepackage[normalem]{ulem}
\usepackage{wrapfig}
\usepackage{multirow}
\usepackage[export]{adjustbox}
\usepackage{stackengine}    
\usepackage{float}    
\usepackage[edges]{forest}
\usepackage{array}
\usepackage{stfloats}
\usepackage[implicit=false]{hyperref}
\hypersetup{
 colorlinks,
 linkcolor=blue,
 urlbordercolor=black,
 pdfborderstyle={/S/U/W 1}
 }

\definecolor{folderbg}{RGB}{124,166,198}
\definecolor{folderborder}{RGB}{110,144,169}
\definecolor{IGNGREEN}{RGB}{153, 211, 142}
\definecolor{TITLES}{RGB}{153, 211, 142}
\definecolor{TITLES_PRE}{RGB}{247, 212, 188}

\newlength\Size
\titleformat{name=\section}[block]
  {\centering\rmfamily\small}
  {}
  {0pt}
  {\colorsection}
\titlespacing*{\section}{0pt}{\baselineskip}{\baselineskip}
\newcommand{\colorsection}[1]{%
  \colorbox{TITLES_PRE}{\parbox{\dimexpr\linewidth-2\fboxsep}{\thesection\ #1}}}

\def\BibTeX{{\rm B\kern-.05em{\sc i\kern-.025em b}\kern-.08em
    T\kern-.1667em\lower.7ex\hbox{E}\kern-.125emX}}
    
\tikzset{%
  folder/.pic={%
    \filldraw [draw=folderborder, top color=folderbg!50, bottom color=folderbg] (-1.05*\Size,0.2\Size+5pt) rectangle ++(.75*\Size,-0.2\Size-5pt);
    \filldraw [draw=folderborder, top color=folderbg!50, bottom color=folderbg] (-1.15*\Size,-\Size) rectangle (1.15*\Size,\Size);},
  file/.pic={%
    \filldraw [draw=folderborder, top color=folderbg!5, bottom color=folderbg!10] (-\Size,.4*\Size+5pt) coordinate (a) |- (\Size,-1.2*\Size) coordinate (b) -- ++(0,1.6*\Size) coordinate (c) -- ++(-5pt,5pt) coordinate (d) -- cycle (d) |- (c) ;},
}

\forestset{%
  declare autowrapped toks={pic me}{},
  pic dir tree/.style={%
    for tree={folder,font=\ttfamily,grow'=0,},
    before typesetting nodes={%
      for tree={edge label+/.option={pic me},},},
  },
  pic me set/.code n args=2{%
    \forestset{%
      #1/.style={inner xsep=2\Size,pic me={pic {#2}},}
    }
  },
  pic me set={directory}{folder},
  pic me set={file}{file},
}

\newenvironment{Tabular}[2][1]
  {\def\arraystretch{#1}\tabular{#2}}
  {\endtabular}

\begin{document}

\twocolumn[{
  \begin{@twocolumnfalse}
    \centering\underline{\makebox[16cm]{\LARGE{FLAIR: French Land cover from Aerospace ImageRy.}}} \\
    \vspace{-0.38cm}
    \begin{figure}[H]
    \centering
    \includegraphics[width=2.052\linewidth]{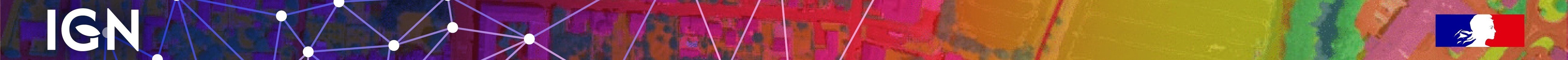}
    \end{figure}
    \centering
    \vspace{-0.3cm}
    \Large{\textit{Challenge FLAIR \#1: }} \\
    \Large{\textit{semantic segmentation and domain adaptation}} \\
    \vspace{+0.4cm}
    \normalsize{Anatol Garioud, Stéphane Peillet, Eva Bookjans, Sébastien Giordano, Boris Wattrelos}    \\
    \vspace{+0.2cm}
    \normalsize{Institut national de l’information géographique et forestière (IGN), France} \\
    \vspace{+0.2cm}
    \centering\normalsize{\textit{ai-challenge@ign.fr}}                  
    \vspace{-0.2cm}
    \begin{figure}[H]
    \centering
    {\setlength{\fboxsep}{0pt}%
    \setlength{\fboxrule}{1.5pt}%
    \fbox{\includegraphics[width=2.04\linewidth]{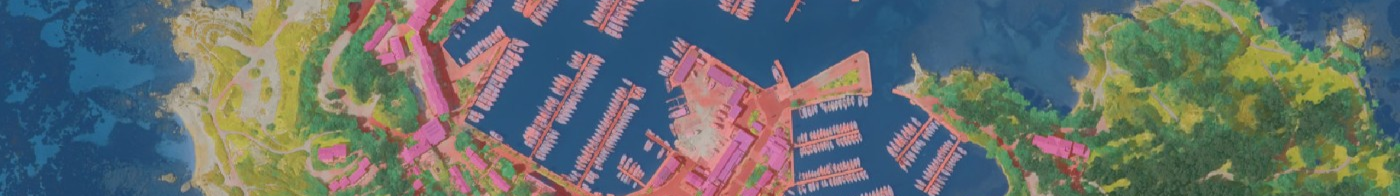}}}
    \end{figure}    
  \end{@twocolumnfalse}
}]

%

\section*{\textbf{Dataset overview}}
\vspace{-0.5cm}
\begin{table}[h!]
\centering
\renewcommand{\arraystretch}{1.5}
\begin{Tabular}[1.5]{|p{8.2cm}|}
\hline \rowcolor[HTML]{C0C0C0} \textbf{Figures}                    \\  \hline
\end{Tabular}
\par\vskip1.2pt
\begin{Tabular}[0.8]{|p{8.2cm}|}
\hline
\rowcolor[HTML]{dbdbdb} \color{black}\ding{212} \color{black} 20,293,091,328 pixels annotated at 0.20\:m spatial resolution \\
\rowcolor[HTML]{dbdbdb} \color{black}\ding{212} \color{black} 77,412 patches (512$\times$512) \\
\rowcolor[HTML]{dbdbdb} \color{black}\ding{212} \color{black} 50 spatio-temporal domains and 950 areas covering 810 km²\\
\rowcolor[HTML]{dbdbdb} \color{black}\ding{212} \color{black} 13 semantic classes (+6 optional ones)        \\[1em] \hline
\end{Tabular}
\end{table}

\vspace{-0.5cm}

\begin{table}[h!]
\centering
\renewcommand{\arraystretch}{1.5}
\begin{Tabular}[1.5]{|p{8.2cm}|}
\hline \rowcolor[HTML]{C0C0C0} \textbf{Structure}                    \\  \hline
\end{Tabular}
\par\vskip1.2pt
\hspace{0.02cm}
\begin{Tabular}[0.8]{|p{8.2cm}|}
\hline 
\rowcolor[HTML]{dbdbdb}  {\footnotesize
\begin{forest}
  pic dir tree, where level=0{}{directory,},
  for tree={ s sep=0.05cm, l sep=1cm, font=\rmfamily }
  [\textbf{Dataset}
    [\textbf{train\textbackslash val}
      [domain \_ year
        [area 
          [img
            [IMG\_ID.tif, file]]
          [msk
            [MSK\_ID.tif, file]]            
            ]
      ]
    ]
    [\textbf{test}
      [domain\_year
        [area
          [img
            [IMG\_ID.tif, file]]    
        ]  
      ]
    ]
    [metadata.json, file]
  ]
\end{forest}}      \\ \hline
\end{Tabular}
\vspace{-4mm}
\end{table}

\section*{\textbf{Context}}
The French National Institute of Geographical and Forest Information (IGN)\cite{IGN} has the mission to document and measure land-cover on French territory and provides referential geographical datasets, including high-resolution aerial images and topographic maps. The monitoring of land-cover plays a crucial role in land management and planning initiatives, which can have significant socio-economic and environmental impact. Together with remote sensing technologies, artificial intelligence (IA) promises to become a powerful tool in determining land-cover and its evolution.

IGN is currently exploring the potential of IA in the production of high-resolution land cover maps. Notably, deep learning methods are employed to obtain a semantic segmentation of aerial images. However, territories as large as France imply heterogeneous contexts: variations in landscapes and image acquisition make it challenging to provide uniform, reliable and accurate results across all of France.

The FLAIR-one dataset presented is part of the dataset currently used at IGN to establish the French national land cover map reference \textit{Occupation du sol à grande échelle} (OCS-GE).


\section*{\textbf{Domain adaptation challenge}}

The aerial surveys covering the French territory at very high spatial resolution raise substantial domain adaptation challenges: \vskip 0.18cm

\textbf{Large-scale coverage:} vast territories the size of a country, such as France, include varied landscapes and climates. Consequently, a single semantic class encompasses very heterogeneous characteristics. For example, agricultural crops grown in Northern France are different, in terms of species and phenology, to those in Southern France. Likewise, forest species composition varies significantly between coastal and mountainous areas. Besides obviously affecting vegetation land cover classes, large-scale discrepancies in land cover characteristics affect almost all semantic classes to some extent: the roofing depends on regional building practices and therefore can be dissimilar; the color of bare soils varies widely depending on the soil type and so on.\vskip 0.18cm

\textbf{Temporal shift:} images within a particular aerial survey are acquired between April and November. Therefore a given area can include multiple seasons. This particularity greatly affects the appearance of the vegetation land cover classes. For instance, wheat crops are pictured at different stages of growth: from bare soil to low and fully developed plants back to bare soil after harvest. Significant seasonal changes also occur in deciduous trees or forested areas.\vskip 0.18cm

\textbf{Multi-sensor acquisition:} different cameras are used for the aerial image acquisitions. This implies different sensors and consequently, different image characteristics. For example, even if the final product is delivered at a specified target spatial resolution (\textit{i.e.} 0.2\:m), depending on the camera, the actual spatial resolution can vary slightly. The same occurs with radiometric information, as the spectral sensitivity (bandwidth, center value, etc.) of the channels (blue, green, red, near-infrared) is different among the used cameras.\vskip 0.18cm

\begin{figure}[htpb]
\centering
\includegraphics[width=1\linewidth]{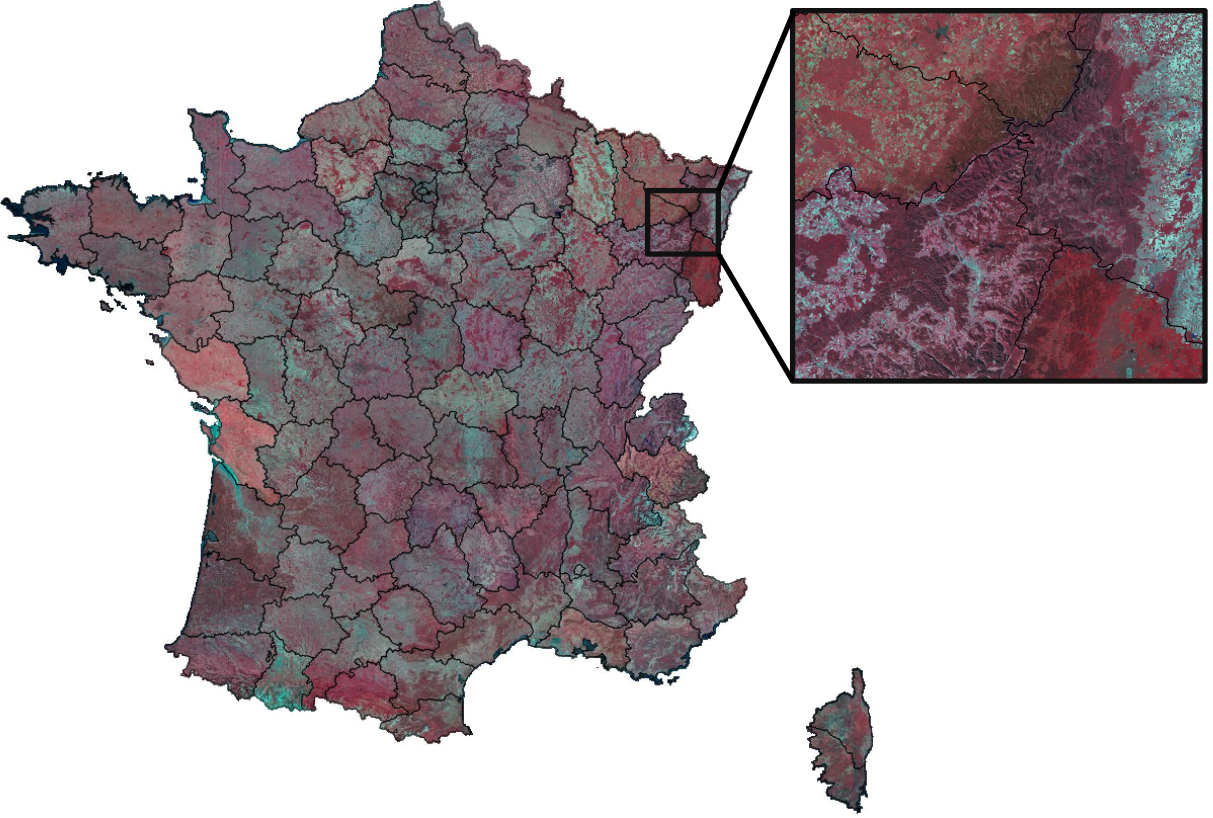}
\caption{Radiometric discrepancies of the aerial images between areas.The 3 channels image displayed is a composite of Near-Infrared, Red and Green spectral information.}
 \label{fig:irc-areas}
\end{figure}

\textbf{Radiometric processing:} within a spatial domain (see next section for a definition), all aerial acquisitions are radiometrically corrected to reduce disparities in sunlight and contrast. Nonetheless, this homogenization is not applied across all the different spatial domains (Figure~\ref{fig:irc-areas}) equally. As opposed to satellite imagery, the pixel intensity in the image channels can therefore not be considered as a physical measure.

In summary, spatial and temporal heterogeneity both in semantic and radiometric terms make it challenging for generalizations of learned models on the FLAIR-one dataset.


\titleformat{name=\section}[block]
  {\centering\rmfamily\small}
  {}
  {0pt}
  {\colorsection}
\titlespacing*{\section}{0pt}{\baselineskip}{\baselineskip}
\renewcommand{\colorsection}[1]{%
  \colorbox{TITLES}{\parbox{\dimexpr\linewidth-2\fboxsep}{\thesection\ #1}}}

\section*{\textbf{Spatial and temporal domains definition}}
\begin{figure}[htpb]
\centering
\includegraphics[width=1\linewidth]{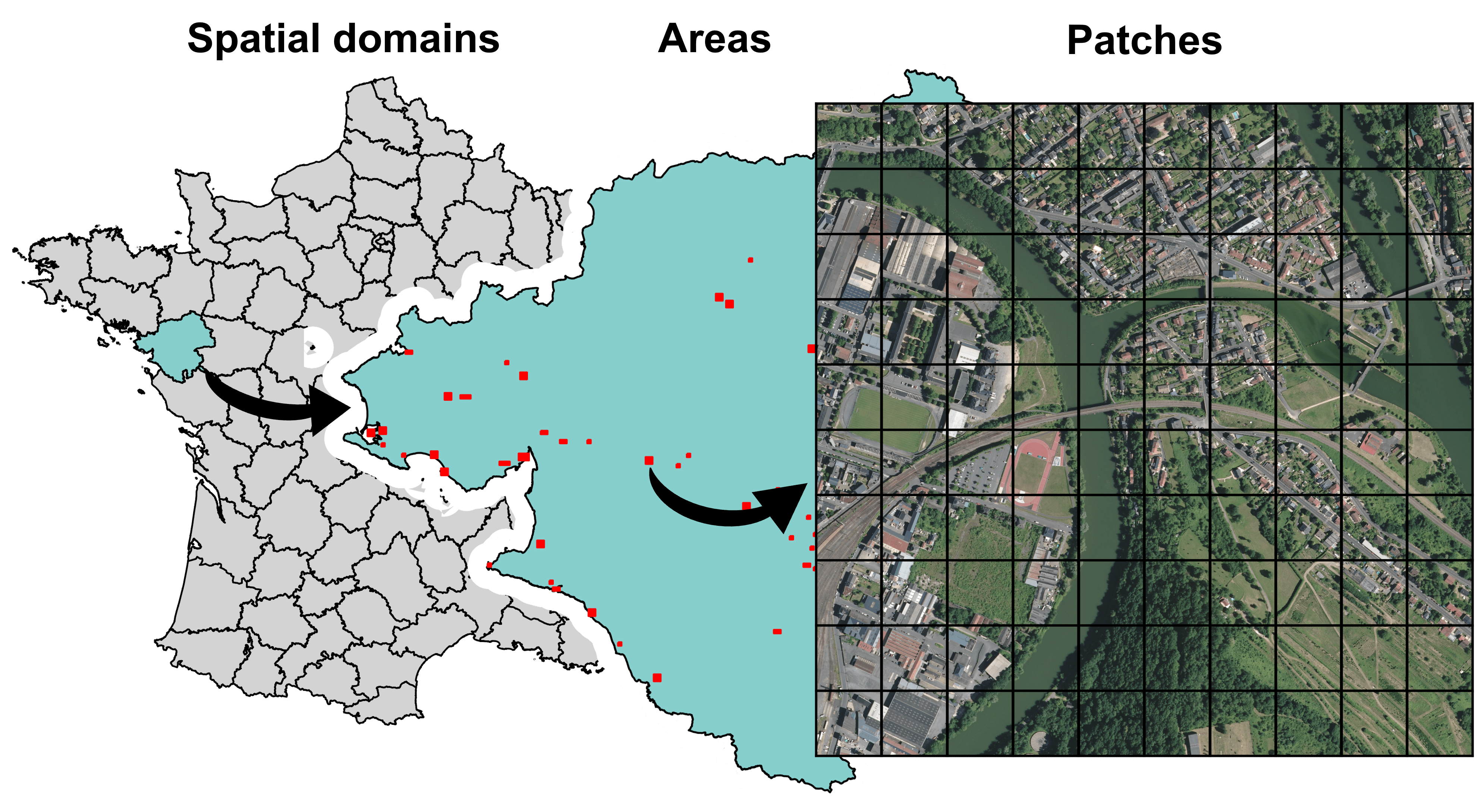}
\caption{Spatial domains, areas and patches.}
\label{fig:spatial_def}
\end{figure}

\textbf{Spatial domains:} a spatial domain is equivalent to a "\textit{département}" and is therefore determined by administrative boundaries. Within each spatial domain, areas of similar sizes have been defined. These areas are subdivided in patches, which are the same size across the dataset (Figure~\ref{fig:spatial_def}). The total number of patches within a spatial domain varies only slightly, ensuring a balanced sampling per spatial domain.\vskip 0.18cm

\noindent \textbf{Temporal domains:} covering large scale territories with aerial imagery takes time. Approximately three years are needed to cover the entire metropolitan French territory. As previously stated, the aerial images are usually acquired between the months of April and November depending on the availability of IGN's planes and on the cloud cover, to ensure that resulting images are cloudless. Consequently, acquisitions within a single spatial domain are sometimes taken on several dates, possibly differing by several months thus encompassing different temporal domains. Furthermore, even within an area of a spatial domain the acquisitions date may vary. 

\section*{\textbf{Dataset extent}}
\begin{figure}[htpb]
\centering\includegraphics[width=0.85\linewidth]{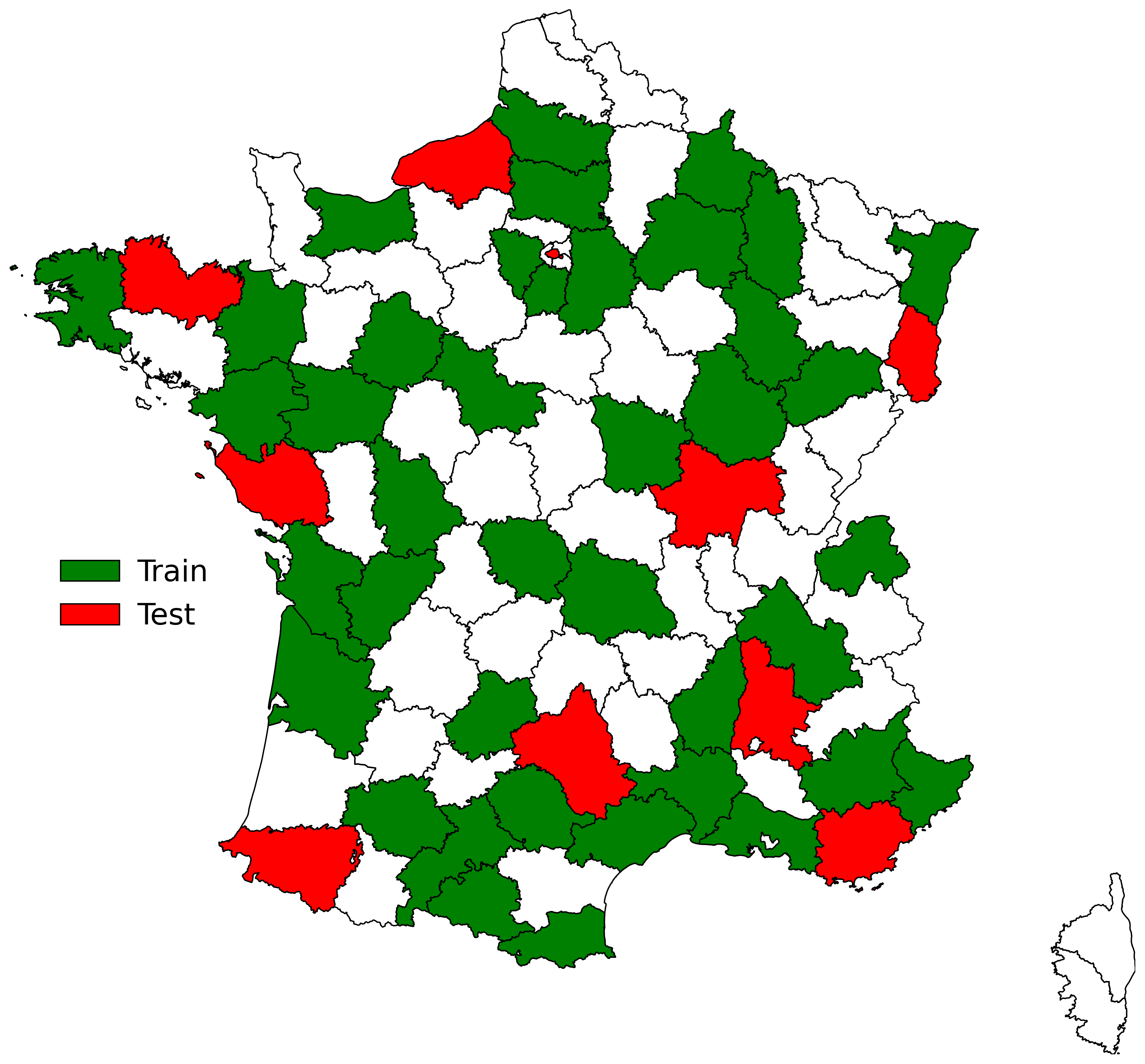}
\caption{The 50 spatial domains of the dataset and the train/test split.}
\label{fig:spatial_dom}
\end{figure}

The dataset includes $50$ spatial domains (Figure~\ref{fig:spatial_dom}) representing the different landscapes and climates of metropolitan France. The train dataset constitutes \nicefrac{4}{5} of the spatial domains (40) while the remaining \nicefrac{1}{5} domains (10) are kept for testing. Within each spatial domain several areas have been selected for a total of approximately 18\:km² resulting in a total of $951$ areas spread across the French metropolitan territory. 

\begin{figure}[htpb]
\centering\includegraphics[width=0.85\linewidth]{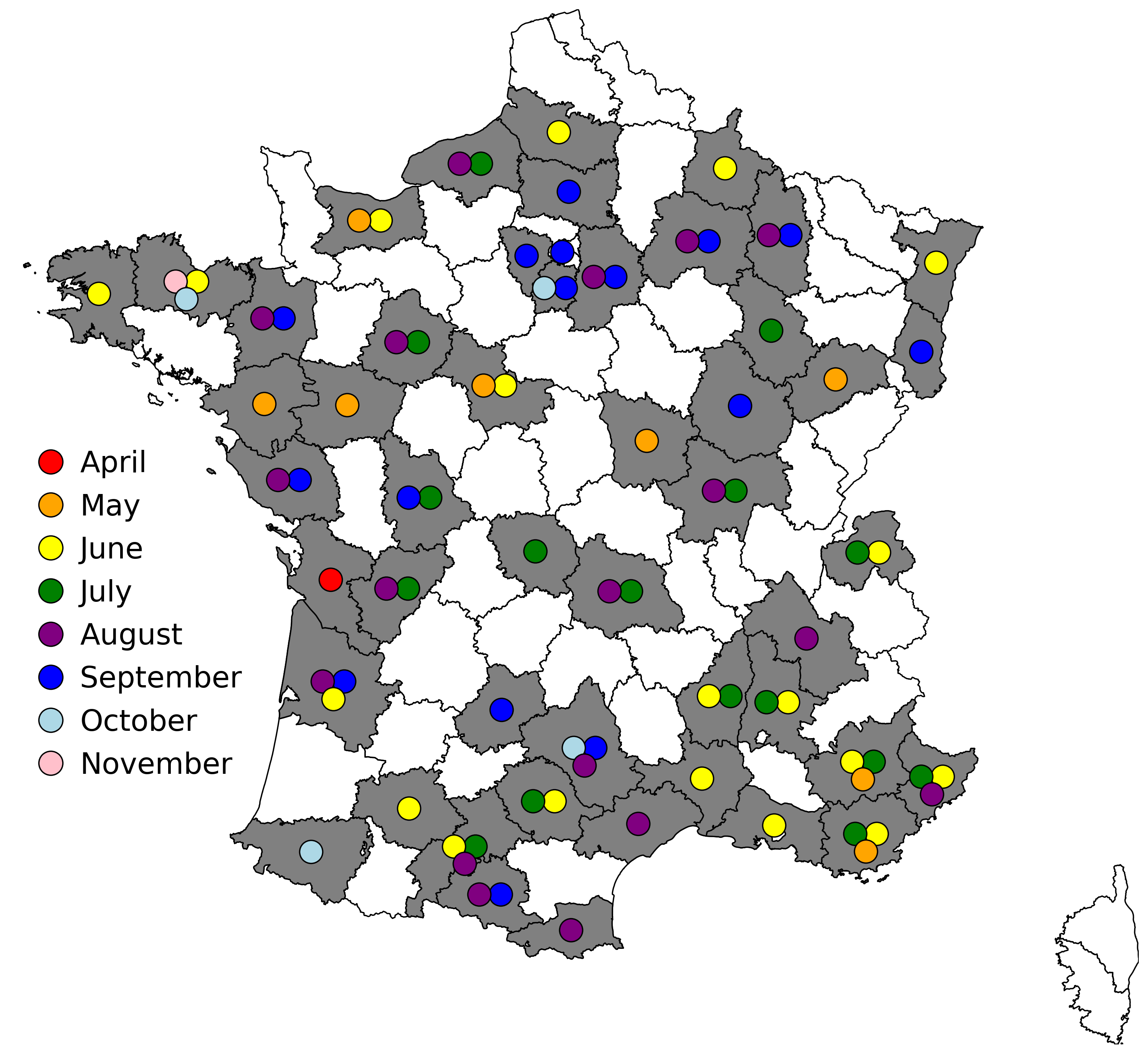}
\caption{Aerial imagery acquisition months per spatial domain.}
\label{fig:tempo_dom}
\end{figure}

The selection of spatial domains takes into account the major landscapes (coastal, urban, agricultural, rural, mountainous,...), the diversity of semantic classes and the variations in radiometry across domains. As a result, the 50 spatial domains can be considered to be representative of the land-cover diversity found in metropolitan France. Additionally, the spatial domains have been selected to include the different temporal domains characteristic of large-scale aerial imagery. An effort was made to balance the acquisitions dates in terms of months (Figure~\ref{fig:tempo_dom}).

A perfect balance, however, is not possible given that certain months have a greater probability to have a cloudless day than others and therefore are naturally more heavily represented.

\section*{\textbf{Data sources}}
An aerial survey is a collection of individual aerial images covering an entire spatial domain, whereby the individual aerial images overlap each other in such a way that 3D information can be extracted. An individual image covers only a small portion of a spatial domain; approximately 3,000 images are needed to cover a medium-sized spatial domain. Each image is characterized by a unique time and date of acquisition and the properties of the camera used.\vskip 0.16cm

Two types of aerial cameras were used in the FLAIR-one dataset containing 50 spatial domains: the Ultracam Eagle M3 camera from Vexcel Imaging and the CAMv2 camera from IGN \cite{souchon2012}. Both types of cameras are composed of 4 multispectral (MS) channels and a panchromatic channel (PAN). The MS channels correspond to 3 spectral bands in the visible part of the electromagnetic spectrum: Blue (B), Green (G), Red (R) and a spectral band in the near infrared (NIR). The PAN channel is a unique channel covering the entire visible spectrum. The PAN channel has a higher spatial resolution than the MS channels. To obtain a similar spatial resolution, pan-sharpening techniques are used combining R, G, B and PAN channels. The NIR channel is artificially oversampled to match the PAN spatial resolution using traditional interpolation techniques. We note that the resolution of the MS channels varies between the two types of cameras used: for the Ultracam Eagle M3 camera pixel, the spatial resolution of the MS channels is lower by a factor of three compared to the PAN channel resolution, whereas for the CAMv2 camera it is lower by a factor of two. As a result, the actual spatial resolution of the MS channels can slightly differ depending on the camera type used. \vskip 0.20cm 

The FLAIR-one dataset is based on 3 data sources: ORTHO HR\textsuperscript{\textregistered} imagery, the Digital Surface Model (DSM) and the RGE ALTI Digital Terrain Model (DTM). The ORTHO HR\textsuperscript{\textregistered} and DSM sources are generated from the aerial survey of the domain, whereas the DTM is computed with auxiliary sources.\vskip 0.16cm

\noindent\textbf{ORTHO HR}\textsuperscript{\textregistered}: the ORTHO HR\textsuperscript{\textregistered} imagery is a mosaic of all the individual images taken during an aerial survey and mapped onto a cartographic coordinate reference system. The individual images are projected to the RGE ALTI DTM, which provides solely the altitude of the ground. Since the height of the buildings and vegetation (\textit{e.g.}, trees) is not taken into consideration, these objects can appear to lean on the ORTHO HR\textsuperscript{\textregistered} images depending on their position in the image. The final ORTHO HR\textsuperscript{\textregistered} product has a spatial resolution of 0.20 m (R, G, B and NIR channels). Additionally, some radiometric processing methods are applied to obtain the final product. First, radiometric equalization methods are applied to the individual images. Then, a global radiometric correction is carried out on the merged images covering an entire spatial domain to provide a more satisfying color balance between channels. Consequently, the shift in radiometry observable between spatial domains (Figure~\ref{fig:irc-areas}) is due to both the date of acquisition and the specific radiometric corrections applied. Therefore, the radiometry of R, G, B and NIR images of the ORTHO HR\textsuperscript{\textregistered}  product cannot be considered as a physical measurement of channel reflectance.\vskip 0.16cm 

\noindent\textbf{Digital Surface Model (DSM)}: the DSM gives the altitude, in meters, for each pixel. Thanks to dense matching techniques, the DSM is derived from the same aerial survey that is used to produce the ORTHO HR\textsuperscript{\textregistered}. This gives the DSM the same spatial resolution as the ORTHO HR\textsuperscript{\textregistered} (0.20\:m), but more importantly prevents temporal shifts and ground cover changes between the two products making them temporally coherent. However, there still are small geometric differences between the two products. Buildings and vegetation objects contained in the DSM do not lean as it is the case in the ORTHO HR\textsuperscript{\textregistered} product, possibly resulting in small shifts, which are more visible for high altitude objects. Moreover, dense matching techniques are applied automatically, potentially introducing noise and artifacts. In particular, radiometrically homogeneous areas (\textit{e.g.}, parts of the aerial image with little texture) tend to lead to locally false and varying 3D information. Apart from these artifacts, the vertical accuracy of the DSM is considered to be twice the spatial resolution (0.4\:m). \vskip 0.16cm

\noindent \textbf{RGE ALTI Digital Terrain Model (DTM):} this product is a national DTM available at a spatial resolution of 1\:m. It is constructed from different sources such as dense matching of aerial images, airborne Lidar or, for mountainous areas, from airborne Synthetic Aperture Radar acquisitions. Depending on the source, the vertical accuracy of the RGE ALTI DTM can vary between 0.3\:m and 7\:m.

\begin{figure*}[htpb]
\centering\includegraphics[width=0.89\linewidth]{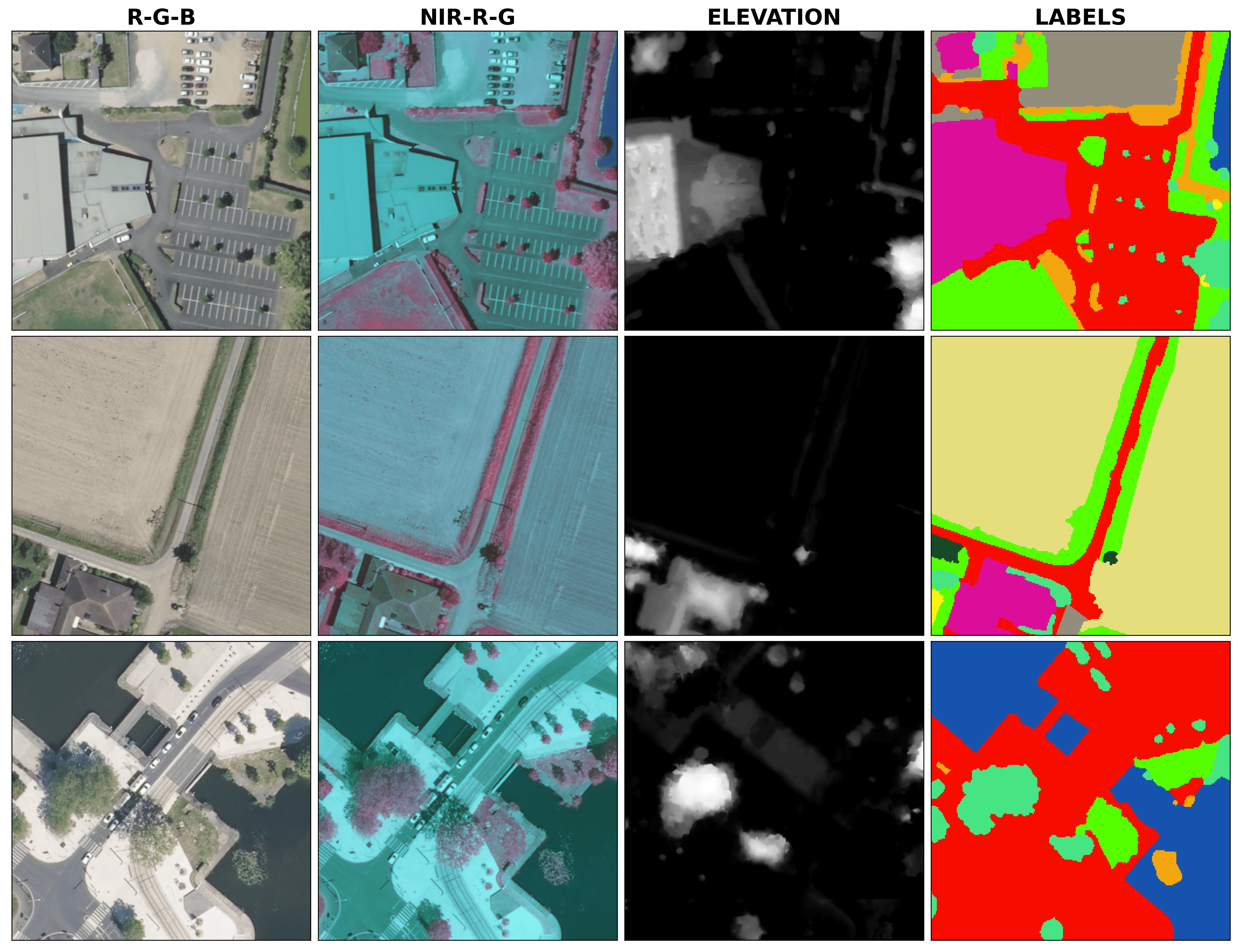}
\caption{Example of input and supervision data: true color composition, near-infrared color composition, elevation band and supervision masks. The data from the first three columns are retrieved from the IMG files while the last column corresponds to the MSK files.}
\label{fig:pex}
\end{figure*}

\section*{\textbf{Data description and naming conventions}}
The dataset is composed of 77,412 patches for a total of 20,319,305,628 pixels covering approximately 810\:km². Each patch is 512$\times$512 pixels corresponding to 10,485.76\:m$^{2}$ on the ground. Preprocessing of the dataset was mainly carried out with the odeon-landcover library \cite{ODEON}.\vskip 0.15cm

\noindent The \textbf{input patches (IMG)} have 5 channels (5$\times$512$\times$512) corresponding to: Blue, Green, Red, Near-Infrared and Elevation channels. The first 4 channels are retrieved from the ORTHO HR\textsuperscript{\textregistered} aerial imagery and are encoded as 8bit unsigned integer datatype. The fifth channel is the difference between the Digital Surface Model (DSM) and the Digital Terrain Model (DTM) providing information on the above-ground height of the imaged objects. For storage optimization reasons, this elevation information is multiplied by a factor of 5 and encoded as a 8bit unsigned integer datatype.\vskip 0.15cm

\noindent The \textbf{supervision patches (MSK)} have 1 channel (1$\times$512$\times$512), describing the semantic class of the pixel (values between 1 and 19) and is encoded as a 8bit unsigned integer datatype.\vskip 0.15cm

\noindent \textbf{Metadata (MTD):} provides per patch additional information: 
\begin{itemize}[label=\ding{212}]
    \setlength\itemsep{0.4em}
    \item the acquisition date (YYYY-MM-DD) and time (HH:MM:SS) of the aerial image, allowing  the integration of temporal information (\textit{e.g.}, seasons).
    \item the geographical coordinates (XY) of the centroid and the mean altitude (Z) of the patch, providing spatial information on the relative location of the patches to each other on a broader scale.
    \item the camera type used for the aerial image acquisition, providing further information on the data.
\end{itemize}\vskip 0.15cm

\noindent The data structure and naming convention is illustrated in the data overview given above. The filename of the data starts with the type of data (IMG or MSK) followed by ‘\_ID’, whereby each pair of IMG-MSK pair has a unique ID (\textit{e.g.}, 000845). IDs from 000001 to 061713 correspond to the train dataset and those from 061714 to 077413 to the test set. Examples of IMG-MSK pairs are provided in Figure~\ref{fig:pex}. Finally, we note that the final two letters of the area folder name indicate the two major types of land cover in the area. The letter U indicates an urban, N a natural, A an agricultural, and F a forest area.

\section*{\textbf{Label production and semantic classes}}
The thematic application of the dataset consists in determining the land cover at the pixel-level, (\textit{i.e.}, measuring how much part of a territory is covered by buildings, forest and so on), enabling land management authorities to better understand the current landscape and its evolution.\\

The supervision data (MSK) is based on photo-interpretation of the ORTHO HR\textsuperscript{\textregistered} aerial imagery and has been manually produced by experts following a call for tenders from IGN. An initial spatial multi-level image segmentation approach using PYRAM \cite{PYRAM} was applied, simplifying the labeling at the cluster level. We note that this segmentation was not necessarily final, but was modified interactively when deemed appropriate. It was specified that movable objects (\textit{e.g.}, cars, boats) are not to be annotated as such, but to be classified as the underlying cover. For example, a car on an asphalt road is labeled as an impervious surface.\\  

\noindent \textbf{Original nomenclature:}  The dataset has an original semantic richness of 18 common land cover classes and a class \textit{other}. The class \textit{other} corresponds to pixels that could not be assigned to one of the 18 well-defined semantic classes with certainty by the photo-interpreters, due to, for example, radiometry issues or visual obstruction caused by the angle of incidence of acquisition.\\

\begin{table}[htpb]
\small
\centering
\setlength{\tabcolsep}{4.9pt}
\renewcommand{\arraystretch}{1.6}
\begin{tabular}{p{1cm}lccr}
 & \textbf{Class}              & \textbf{MSK}       & \textbf{Pixels}   & \textbf{\%} \\ \hline
\cellcolor[HTML]{db0e9a} & building           & 1                & 1,670,300,028 & 8.23       \\
\cellcolor[HTML]{938e7b} & pervious surface   & 2                & 1,636,681,162 & 8.07       \\
\cellcolor[HTML]{f80c00} & impervious surface & 3                & 2,836,695,330 & 13.98      \\
\cellcolor[HTML]{a97101} & bare soil          & 4                & 741,261,583   & 3.65       \\
\cellcolor[HTML]{1553ae} & water              & 5                & 1,034,960,677 & 5.10       \\
\cellcolor[HTML]{194a26} & coniferous         & 6                & 540,883,511   & 2.67       \\
\cellcolor[HTML]{46e483} & deciduous          & 7                & 3,061,129,298 & 15.08      \\
\cellcolor[HTML]{f3a60d} & brushwood          & 8                & 1,408,504,416 & 6.94       \\
\cellcolor[HTML]{660082} & vineyard           & 9                & 665,099,230   & 3.28       \\
\cellcolor[HTML]{55ff00} & herbaceous vegetation          & 10   & 3,798,549,882 & 18.72      \\
\cellcolor[HTML]{fff30d} & agricultural land               & 11  & 2,061,788,310 & 10.16      \\
\cellcolor[HTML]{e4df7c} & plowed land        & 12               & 720,090,325   & 3.55      \\
\cellcolor[HTML]{000000} & other     & \textbf{\textgreater{}13} & 117,147,576   & 0.58       \\ \hline
\end{tabular}
\caption{Semantic classes of the main nomenclature of the flair-one dataset.}
\label{tab:nom}
\end{table}

\begin{table}[htpb]
\renewcommand{\arraystretch}{1.3}
\colorbox[HTML]{FAEFE4}{
\begin{tabular}{p{8.4cm}}
\multicolumn{1}{c}{\normalsize{\textbf{\underline{Class description}}}} \\ \\[-1.7ex]
\textit{Note: as previously stated, semantic classes are assigned on the cluster level. In a given aerial image, only observable objects are labeled, whereby temporal aspects are not taken into consideration.}\\ \\[-1.6ex]

\textbf{Anthropized surfaces without vegetation (1, 2, 3, 13 and 18)} \\ \hline
\multicolumn{1}{|p{8.4cm}|}{\textbf{\textit{Class 1 – building}} includes not only buildings, but also other type of constructions such as towers, agricultural silos, water towers and dams. Greenhouses (class 18) are an exception.}                              \\
\multicolumn{1}{|p{8.4cm}|}{\textbf{\textit{Class 2 – pervious surface}} defined as man-made bare soils covered with mineral materials (\textit{e.g.} gravel, loose stones) and considered to be pervious. It includes pervious transport networks (\textit{e.g.} gravel pathways, railways), quarries, landfills, building sites and coastal ripraps.}\\
\multicolumn{1}{|p{8.4cm}|}{\textbf{\textit{Class 3 – impervious surface}} is defined as man-made bare soils that are impervious due to their building materials (e.g. concrete, asphalt, cobblestones). It includes roadways, parking lots, and certain types of sports fields.}                                   \\
\multicolumn{1}{|p{8.4cm}|}{\textbf{\textit{Class 13 – swimming pool}} is defined as man-made artificial (open-air) swimming pools. It is not included in class 5 (water).} \\
\multicolumn{1}{|p{8.4cm}|}{\textbf{\textit{Class 18 – greenhouse}} although it can be considered as a building, is given a distinct labeled. Greenhouses are a class of their own and are not part of class 1.} \\ \hline \\[-1.4ex] 

\textbf{Natural areas without vegetation (4, 5 and 14)}\\\hline                                                            
\multicolumn{1}{|p{8.4cm}|}{\textbf{\textit{Class 4 – bare soil}} defined as natural permanently bare soils. These natural soils remain without vegetation throughout the year and generally are covered with sand, pebbles, rocks or stones. Examples of natural bare soils are frequently found in coastal, mountainous and forested areas.}\\                                                             
\multicolumn{1}{|p{8.4cm}|}{\textbf{\textit{Class 5 – water}} is defined as areas covered by water, such as sea, rivers, lakes and ponds. An exception are swimming pools (class 13).}\\                                                             
\multicolumn{1}{|p{8.4cm}|}{\textbf{\textit{Class 14 – snow}} refers to surfaces covered by snow. It is an extremely rare class as the images are taken in the summertime and only very few regions in France are covered with snow year-round.}\\ \hline \\[-1.4ex]     

\textbf{Woody natural vegetation surfaces (6, 7, 8, 15, 16 and 17)}\\\hline
\multicolumn{1}{|p{8.4cm}|}{\textbf{\textit{Class 6 – coniferous}}, is defined as trees identifiable as coniferous (pines, firs, cedars, cypress trees, ...) and taller than 5\:m.}\\
\multicolumn{1}{|p{8.4cm}|}{\textbf{\textit{Class 7 – deciduous}} is defined as trees identifiable as deciduous (oaks, beeches, birches, chestnuts, poplars, ...) and taller than 5\:m.}\\
\multicolumn{1}{|p{8.4cm}|}{\textbf{\textit{Class 8 – brushwood}} refers to natural woody surfaces with a vegetation less than 5\:m high. It includes short and young trees, brushwood, shrublands, mountain moors and abandoned agricultural lands.}\\
\multicolumn{1}{|p{8.4cm}|}{\textbf{\textit{Class 15 – clear-cut}}, is defined as forest areas, in which the trees have been cut down and harvested.}\\
\multicolumn{1}{|p{8.4cm}|}{\textbf{\textit{Class 16 – ligneous}} is an extremely rare class used to describe forest areas with a homogeneous representation of either coniferous or deciduous trees.}\\
\multicolumn{1}{|p{8.4cm}|}{\textbf{\textit{Class 17 – mixed}} is an extremely rare class used to describe forest areas with heterogeneous trees for which the types of trees (coniferous/ deciduous) cannot be determined with sufficient certainty.}\\\hline \\[-1.4ex] 

\textbf{Agricultural surfaces (9, 11 and 12)}\\\hline
\multicolumn{1}{|p{8.4cm}|}{\textbf{\textit{Class 9 – vineyard}} despite being an agricultural use of the land, are assigned a class apart, a reason being their rather distinctive land cover characteristics.}\\
\multicolumn{1}{|p{8.4cm}|}{\textbf{\textit{Class 11 – agricultural land}} encompasses various different agricultural classes. For example, besides major crops, it also includes permanent and temporary grasslands with agricultural use. Vineyards (class 9) are not included in this class.}\\
\multicolumn{1}{|p{8.4cm}|}{\textbf{\textit{Class 12 – plowed land}} is defined as agricultural land with no visible vegetation (\textit{e.g.} recently plowed and freshly harvested land).}\\\hline \\[-1.4ex] 

\textbf{Herbaceous surfaces (10)}\\\hline
\multicolumn{1}{|p{8.4cm}|}{\textbf{\textit{Class 10 – herbaceous vegetation}} defines herbaceous surfaces that are not intensively exploited for agriculture purposes. This class includes ornamental lawns (e.g. gardens, public parks), recreational fields (\textit{e.g.} used for sport), natural herbaceous areas in forested or mountainous areas, non-cultivated grass in agricultural areas or along transportation networks.}\\\hline

\end{tabular}}
\end{table}

\noindent \textbf{Reduced baseline nomenclature:}  Among the 18 classes, some are very strongly under-represented. We therefore propose to re-group the following 6 least common classes together with the original class \textit{other} (value 19): swimming pool (value 13); snow (value 14); clear cut (value 15); mixed (value 16); ligneous (value 17); greenhouse (value 18). \textbf{Users of the dataset are encouraged to use this reduced nomenclature instead of the original one}, in order to avoid being confronted with a very imbalanced segmentation task and to instead focus on the domain adaptation challenge. The baseline code made available includes an automatic remapping of the MSK files during dataloading. If independent code is being developed, this reduction of classes should be integrated.\vskip 0.18cm

It was nonetheless decided to share the MSK files with the full nomenclature (19 classes) to allow a full exploration of the available information. The proposed reduced nomenclature used for the benchmark has 13 land cover classes, presented in Table~\ref{tab:nom}.

\section*{\textbf{Semantic class frequency}}
\begin{figure}[htpb]
\centering\includegraphics[width=1\linewidth]{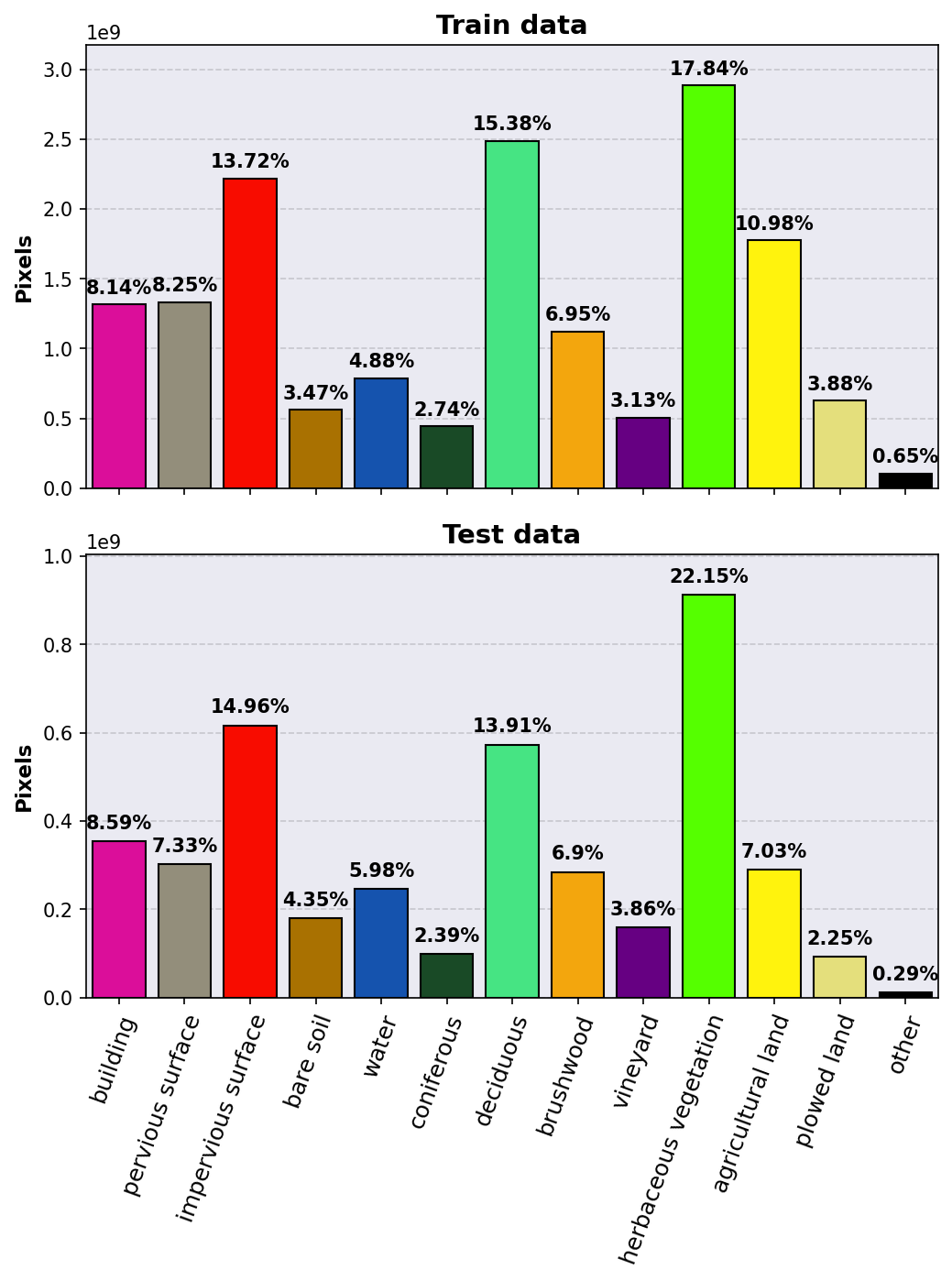}
\caption{Class distribution of the train dataset (\textit{top}) and test dataset (\textit{bottom}).}
\label{fig:class_freq_train}
\end{figure}

The class distribution of the train and test datasets of FLAIR-one are presented in Figure~\ref{fig:class_freq_train} in percentages. 

The train and test datasets have similar semantic class distributions. In both cases, herbaceous vegetation, deciduous trees and impervious surfaces dominate in contrast to bare soil, coniferous, vineyard and plowed land, which in total account for less than 5\% of the landcover in the datasets. It should be noted that the class \textit{other} only represents 0.65\% of the train dataset and 0.29\% for the test dataset, even after regrouping the 6 least common classes of the full nomenclature with the class \textit{other}.

\begin{figure*}[htpb]
\centering\includegraphics[width=1\linewidth]{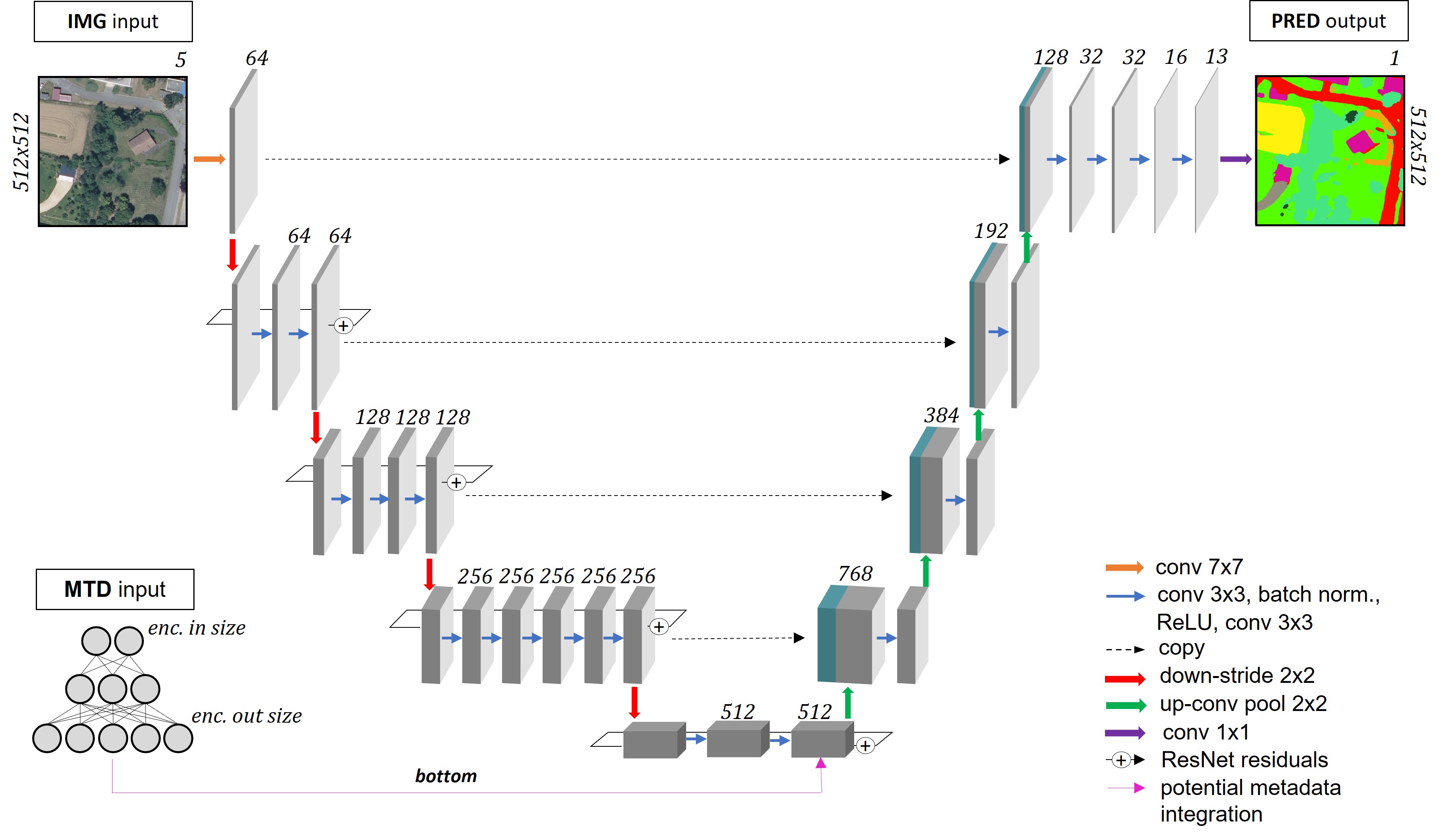}
\caption{U-Net architecture used for the baseline, corresponding to the one implemented in the \textit{segmentation-models-pytorch} library. \textbf{IMG} = input image; \textbf{MTD} = input metadata; \textbf{PRED} = prediction output. One potential and traditional approach to integrate the metadata would be to add a Multi-layer Perceptron for encoding and add the output to the output of the last layer of the encoder or as additional band to the IMG input.}
\label{fig:unet_meta}
\end{figure*}

\section*{\textbf{Benchmark architecture}}
The PyTorch lightning \cite{PNING} framework is used to implement the baseline architecture. We choose a U-Net architecture \cite{U-Net} with a ResNet34 encoder backbone (pre-trained) for a total of $\approx$\:24.4:M parameters and rely on the implementation available in the \textit{segmentation-models-pytorch} library. The architecture employed is illustrated in Figure~\ref{fig:unet_meta}.\\

The Cross Entropy loss (CE) is used as the criterion:
\begin{equation*}
    \mathcal{L_{CE}} = -\sum_{i=1}^nt_{i}\log(p_{i})
\end{equation*}
where $t_{i}$ is the MSK label and $p_{i}$ the Softmax probability for the i$^{th}$ class. Since the class \textit{other} does not contain any information relevant to the segmentation task, we use a class weighting strategy setting its class weight to 0 while the weights of well-defined relevant classes are set to 1.\\

A stochastic gradient descent optimization technique is used during training. A learning rate of $0.02$ is used with a reduction strategy (ReduceLROnPlateau) having a patience set to 10. We train the models with the maximum number of epochs set to 200, but implement an early stopping strategy with a patience of 30 epochs. A batch size of $5$ is used for the baselines.\\

To ensure the reproducibility of the results, we fix the global seed (random, numpy and torch) to \textit{2022}. From the 40 domains of the train dataset, 8 are selected for validation. The following spatial domain split of the dataset is used for the baseline :\vskip 0.2cm

\begin{table}[htpb]
    \centering
    \setlength{\tabcolsep}{3.5pt}
     \renewcommand{\arraystretch}{2.2}
    \begin{tabular}{|lp{5cm}|}
        \hline
       \textbf{\color[HTML]{c7254e}\colorbox[HTML]{f9f2f4}{\texttt{TRAIN}: }}  &  D006, D007, D008, D009, D013, D016, D017, D021, D023, D030, D032, D033, D034, D035, D038, D041, D044, D046, D049, D051, D052, D055, D060, D063, D070, D072, D074, D078, D080, D081, D086, D091\\
       \textbf{\color[HTML]{c7254e}\colorbox[HTML]{f9f2f4}{\texttt{VALIDATION}: }}     & D004, D014, D029, D031, D058, D066, D067, D077 \\
       \textbf{\color[HTML]{c7254e}\colorbox[HTML]{f9f2f4}{\texttt{TEST}: }} & D012, D022, D026, D064, D068, D071, D075, D076, D083, D085\\\hline
    \end{tabular}
    \label{tab:split}
\end{table}\vskip 0.2cm 

\textbf{Users of the dataset intending to report their results to the benchmark must use the same test dataset.}\\

The computational resources used for training are ten NVIDIA Tesla V100 GPUs with 32\:GB memory located on a High Performance Computing (HPC) cluster. The distributed data parallel (ddp) strategy available in PyTorch lightning was used to take advantage of multi-GPU data distribution and computation. The training time with such a configuration is less than 4 hours for the full FLAIR-one dataset.\\ 

Concerning the exploitation of metadata, simple approaches have been tested. The strategies explored have a first step of metadata encoding: positional encoding of spatial and temporal information and one-hot-encoding (OHE) for the camera type. A shallow Multi-layer Perceptron (MLP) with dropout (probability of 0.4) and ReLU activation is then defined to jointly encode the metadata and to provide a specified output size. Subsequently, multiple different integration strategies with the current ResNet34/U-Net segmentation architecture are possible. For example, a commonly employed strategy (depicted as '\textit{bottom}' in Figure~\ref{fig:unet_meta}) consists in matching the MLP output size to the output size of the last layer of the ResNet34 encoder. The two vectors (encoded metadata and encoded images) can then be added and fed into the first layer of the architecture's decoder. Strategies following a similar approaches that add the MLP encoded output at different positions in the architecture's encoder or decoder parts (\textit{e.g.}, after the first input convolution layer, with the last decoder layer, or even added as a sixth channel to the input image) are possible.\\

Commonly used image data augmentation techniques are also tested using the \textit{Albumentation} library. By introducing variance in the dataset, image data augmentation helps to prevent overfitting and provides trained models enhanced generalization capabilities. A wide range of image data augmentation exists, mainly based on geometric (flip, crop, resize, random affine transform...) or color space (brightness, contrast, saturation,...) transformations. For the baseline, only geometric transformations are explored. Vertical and horizontal flips, and random rotations of 0, 90, 180 or 270 degrees are tested.

\begin{table*}[!b]
\small
\centering
\setlength{\tabcolsep}{8pt}
\renewcommand{\arraystretch}{1.4}
\begin{tabular}{lccccc}
      & \textbf{metadata} & \textbf{data augmentation} & \textbf{parameters (M)} & \textbf{epoch saved} & \textbf{mIoU}  \\\hline
\rowcolor[HTML]{D6D5D4}baseline &  \textcolor{red}{\ding{55}} &  \textcolor{red}{\ding{55}}  & 24.4 & 23 & \textbf{0.5443$\pm$0.0014} \\\hline
baseline + \textit{bottom} &    \textcolor{green}{\ding{52}} &  \textcolor{red}{\ding{55}}  & 24.6 & 20 & \textbf{0.5536$\pm$0.0014} \\
baseline + \textit{augmentation} &    \textcolor{red}{\ding{55}} &  \textcolor{green}{\ding{52}}  & 24.4 & 55 & \textbf{0.5335$\pm$0.0038} \\
baseline + \textit{bottom} + \textit{augmentation} &  \textcolor{green}{\ding{52}} &  \textcolor{green}{\ding{52}}  & 24.6 & 40 & \textbf{0.5570$\pm$0.0027} \\\hline
\end{tabular}
\caption{Baseline results of ResNet34/U-Net architecture with different strategies on the FLAIR-one test set. Results are averages of 5 runs of each configuration.}
\label{tab:baseline}
\end{table*}

\section*{\textbf{Benchmark metric}}
To evaluate the semantic segmentation task, first, confusion matrices for each test patch are calculated between the ground truth masks (MSK) and the model predictions (PRED). The resulting test patch confusion matrices are then summed obtaining a single confusion matrix describing the test dataset. Subsequently, the Intersection-over-Union (IoU, or Jaccard Index) metric, of each semantic class, is calculated from this summed confusion matrix:\vskip 0.05cm
\begin{equation*}
    IoU = \frac{|U \cap V|}{|U \cup V|} = \frac{TP}{TP+FP+FN} 
\end{equation*}\vskip 0.05cm
where U denotes the intersection, V the union, TP = true positives, FP = false positives and FN = false negatives.

To obtain a final and single metric allowing the different results to be assessed and ranked, the mean Intersection-over-Union (\textbf{mIoU}) defined as the average of the per-class IoU is calculated. \textbf{However, since the class \textit{other} is ill-defined and equivalent to void and therefore should not affect the results, its IoU is not taken into account for the mIoU calculation.} The final mIoU is thus the average of the 12 remaining IoUs obtained for the test dataset. \\

\begin{figure}[htpb]
\centering\includegraphics[width=0.75\linewidth]{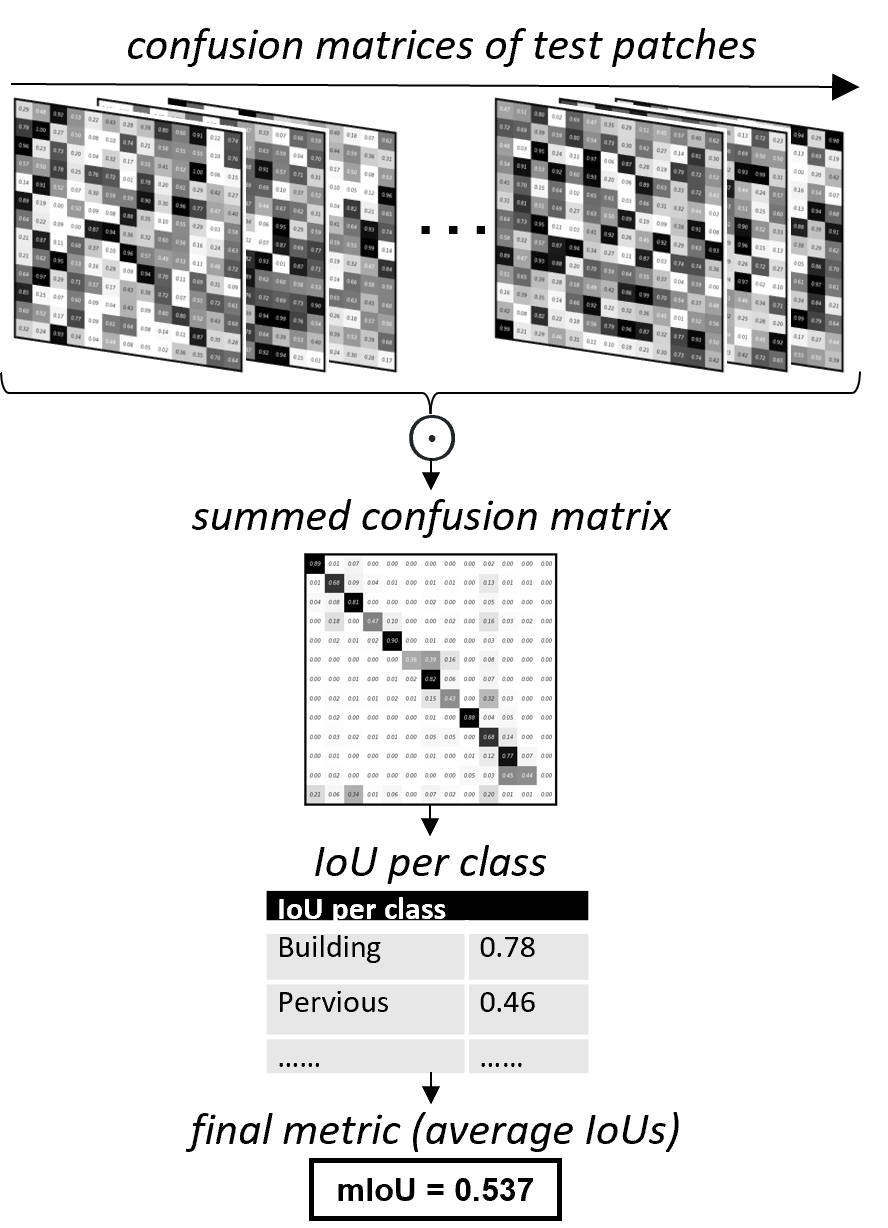}
\caption{Confusion matrices derived from each patch are summed before the IoU calculation per semantic class. The final metric (mIoU) is provided by the averaging of the IoUs.}
\label{fig:eval}
\end{figure}

The workflow of the mIoU calculation is illustrated in Figure~\ref{fig:eval}.

\section*{\textbf{Benchmark results}}
The results obtained using $61,712$ patches for training, and testing on the remaining $15,700$ patches of the FLAIR-one dataset are reported in Table~\ref{tab:baseline}. The given results are average and their standard deviation of 5 mIoU scores obtained for 5 runs in a given configuration.

Besides the baseline result, we also report in Table~\ref{tab:baseline} indicative results about metadata integration and data augmentation. The '\textit{bottom}' strategy refers to the adding the MLP encoded metadata to the last layer of the architecture's encoder. The '\textit{augmentation}' strategy uses the three geometrical augmentations previously described with a probability of 0.5.

We note that the results are not improved with the current use of metadata or data augmentation. Note that potential of metadata integration or image data augmentation has not fully been explored. Opportunities for improvement can be pursued: for example, the encoding methods of the different modalities and their integration into the learning, or strategically applying the image data augmentations by taking into account the semantic class frequencies.\\

\begin{figure}[htpb]
\centering\includegraphics[width=1\linewidth]{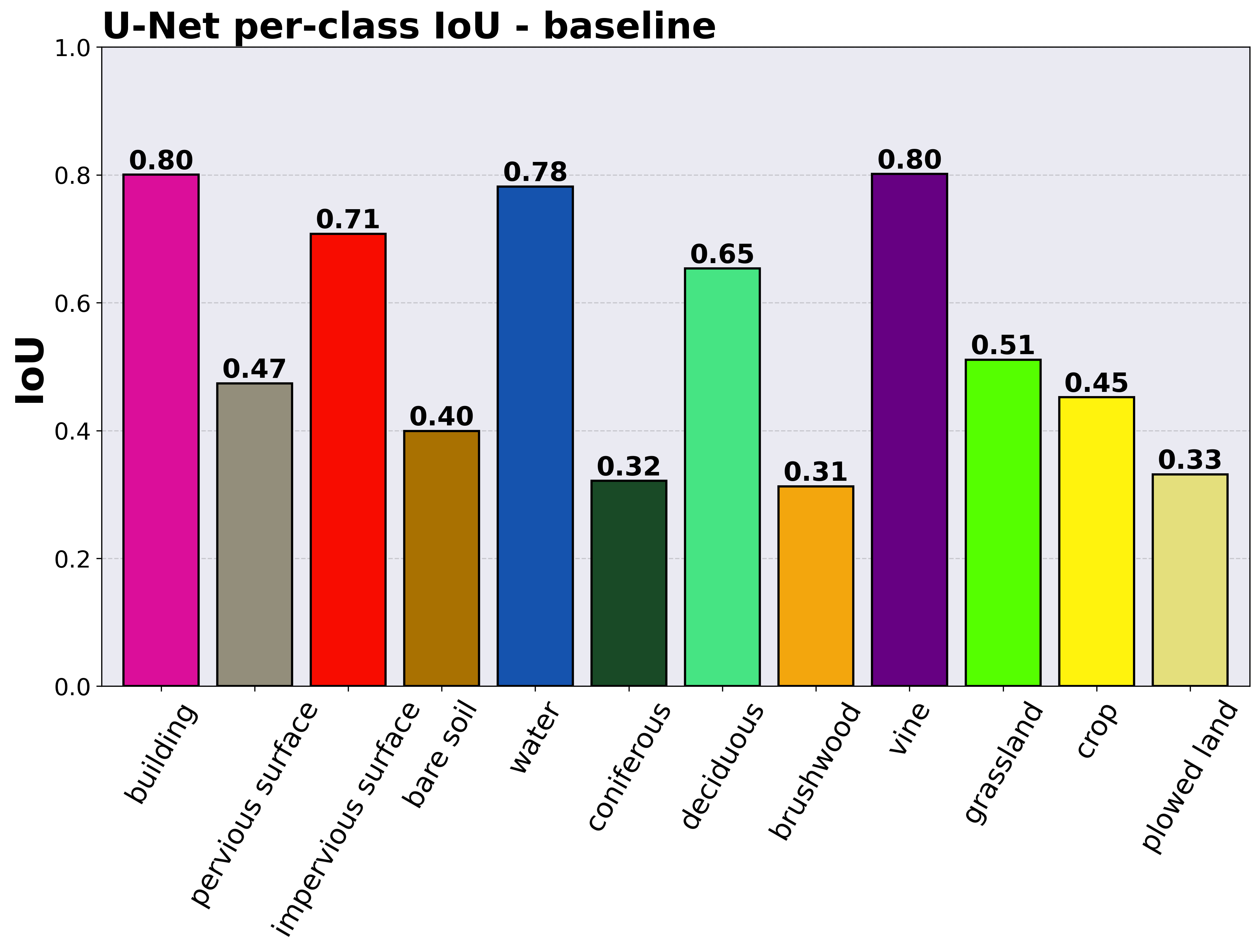}
\caption{Per-class IoU obtained from the baseline architecture on the FLAIR-one dataset.}
\label{fig:results}
\end{figure}

Detailed per-class IoU results for the baseline (without metadata integration or data augmentations) are illustrated in Figure~\ref{fig:results}. 

We observe highly accurate results with IoU $\approx$\:0.8 for three semantic classes (considering the large and heterogeneous dataset used): buildings, water and vineyards. In contrast, other classes exhibit relatively poor results $< 0.4$, such as bare soil, coniferous, brushwood and plowed land. One explanation for these poor results could be due to the relatively low frequency of these classes in the dataset. However, the vineyard class less frequent in the dataset that bare soil, is detected with a significantly higher accuracy.

\begin{figure}[htpb]
\centering\includegraphics[width=1\linewidth]{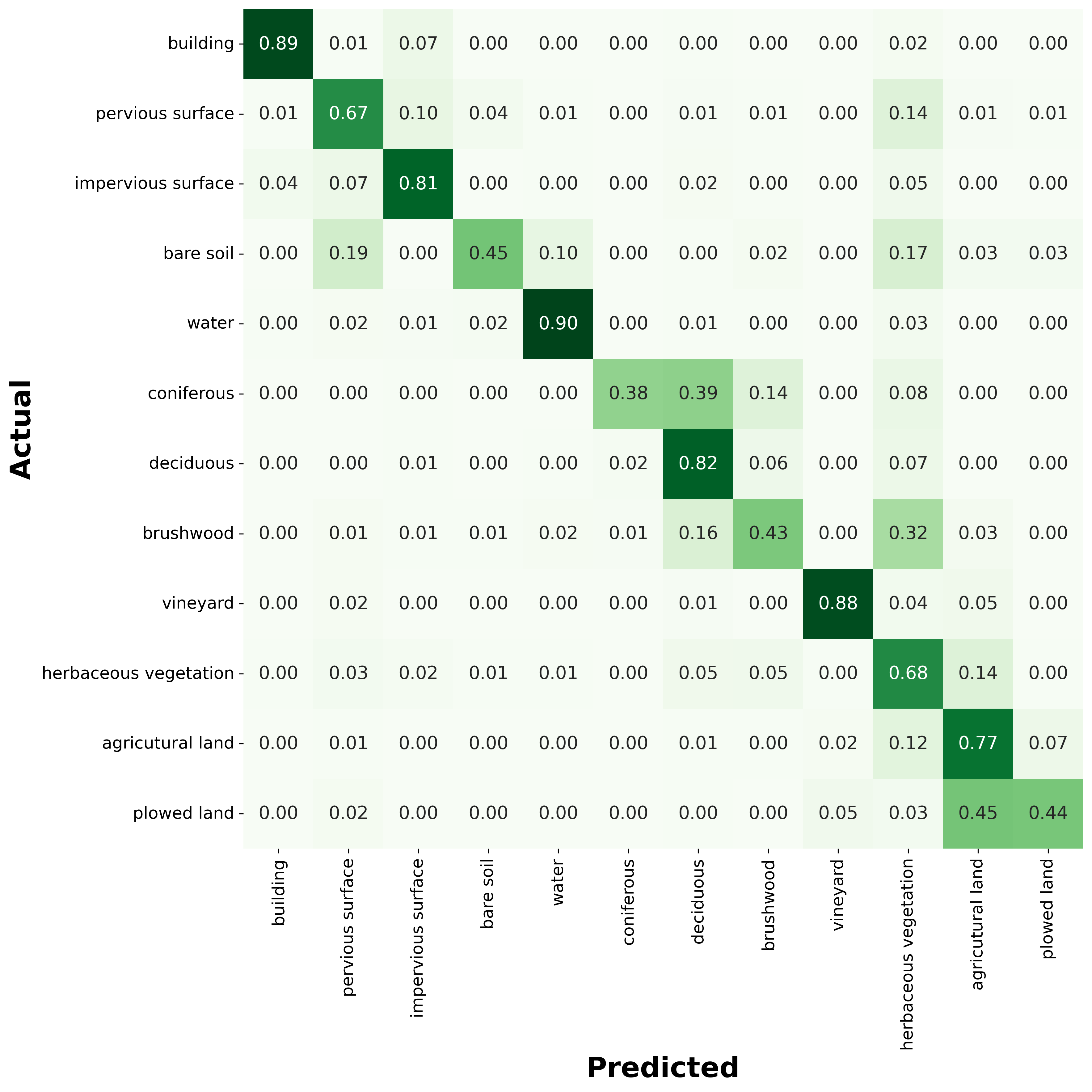}
\caption{Baseline confusion matrix of the test dataset normalized by rows.}
\label{fig:confmat}
\end{figure}

Figure~\ref{fig:confmat} provides the baseline (without metadata integration or data augmentation) confusion matrix of the test dataset, normalized by rows. The confusion matrix shows that inter-class confusions are mainly found between semantically close classes, \textit{e.g.} between herbaceous vegetation and agricultural land, bare soil and pervious surfaces, and coniferous and deciduous.

\vskip 0.6cm

\begin{figure*}[!hb]
\centering
\setlength{\tabcolsep}{0.5pt}
\begin{tabular}{cc}
  \includegraphics[width=0.498\linewidth]{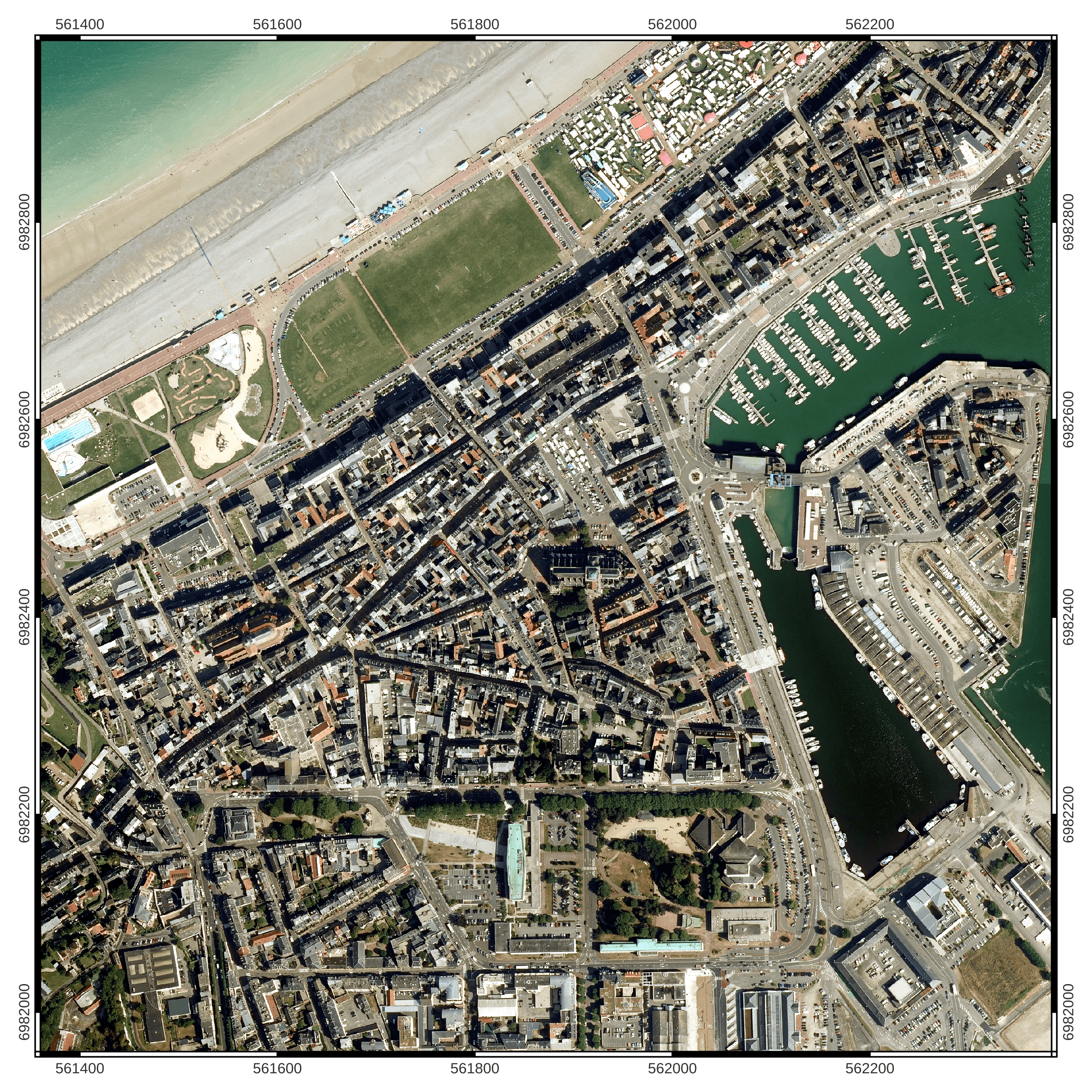} & \includegraphics[width=0.498\linewidth]{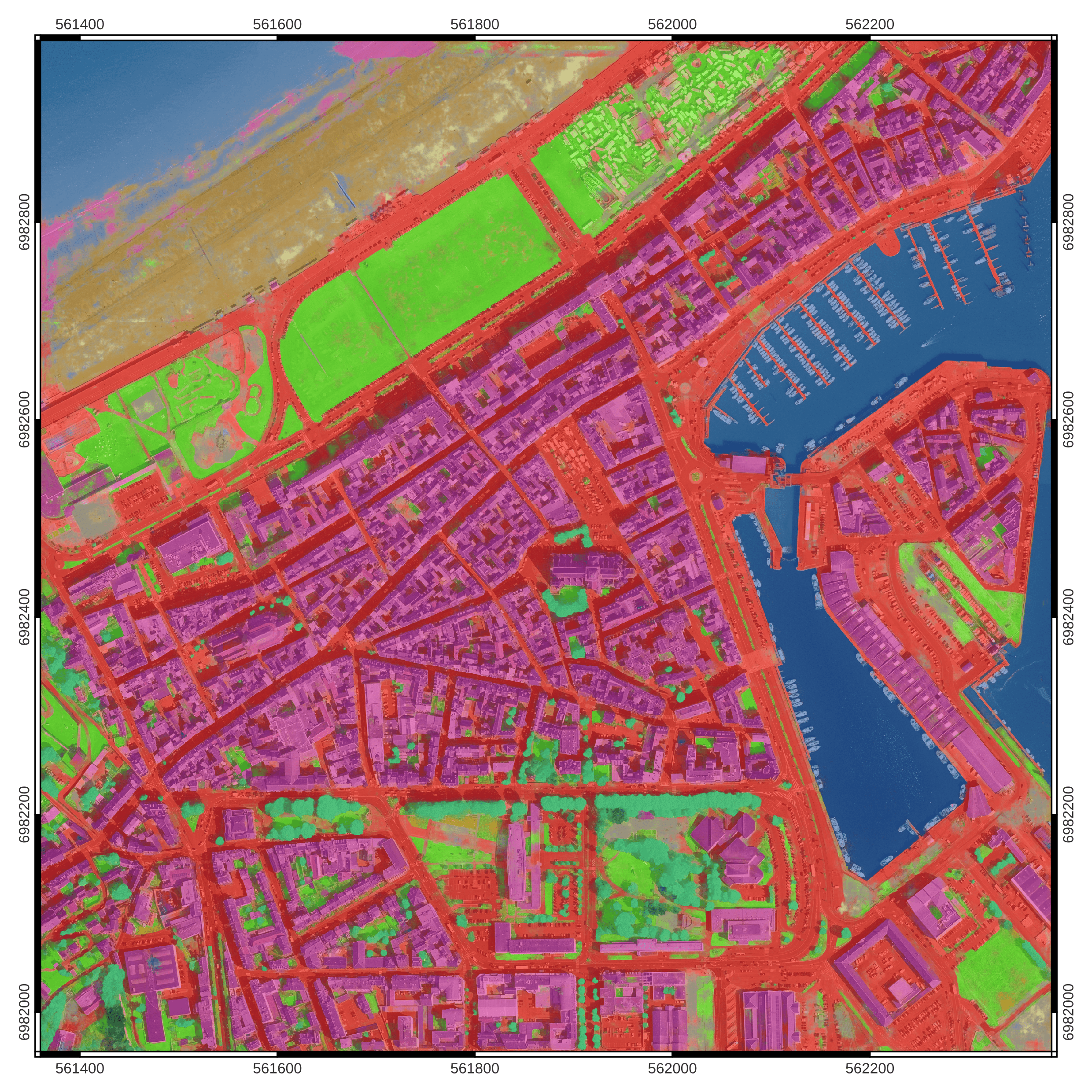} 
\end{tabular}
\caption{Example of a semantic segmentation result using the baseline model.}
\label{fig:example}
\end{figure*}

Finally, Figure~\ref{fig:example} provides an example of a semantic segmentation of an urban and coastal area in the D076 spatial domain, obtained with the baseline trained model.

\bibliographystyle{unsrt}
\bibliography{BIB.bib}

\section*{\textbf{Acknowledgment}}
This work was performed using HPC/AI resources from GENCI-IDRIS (Grant 2022-A0131013803).

\section*{\textbf{Data access}}
The dataset and baseline codes are available at:\\ 
\url{https://ignf.github.io/FLAIR/}

\end{document}